# A deep learning based solution for construction equipment detection: from development to deployment

Saeed Arabi*, Arya Haghighat, Anuj Sharma

Department of Civil, Construction and Environmental Engineering Iowa State University, Ames, IA 50011

**Abstract:** *This paper aims at providing researchers and engineering professionals with a practical and comprehensive deep learning based solution to detect construction equipment from the very first step of its development to the last one which is deployment. This paper focuses on the last step of deployment. The first phase of solution development, involved data preparation, model selection, model training, and model evaluation. The second phase of the study comprises of model optimization, application specific embedded system selection, and economic analysis. Several embedded systems were proposed and compared. The review of the results confirms superior real-time performance of the solutions with a consistent above 90% rate of accuracy. The current study validates the practicality of deep learning based object detection solutions for construction scenarios. Moreover, the detailed knowledge, presented in this study, can be employed for several purposes such as, safety monitoring, productivity assessments, and managerial decisions.*

## 1 INTRODUCTION

Over the years, several methods have been developed by researchers and engineers to monitor civil infrastructure and evaluate their performance. While conventional sensor-based methods are still popular and effective (Amezquita-Sanchez et al., 2018; Amezquita-Sanchez & Adeli, 2016; Arabi et al., 2017, 2018, 2019; Arabi 2018; Constantinescu et al. 2018), recently new vision based methods and solutions, such as deep learning based computer vision solutions (LeCun et al., 2015), have caught attention in different research areas of the civil and infrastructure engineering. Although since decades the main building block of deep learning, i.e. neural networks, has been utilized by researchers (Adeli, 2001), only recently deep learning showed major breakthroughs due to the availability of affordable computing hardware, i.e. Graphics Processing Units (GPUs), as well as large scale datasets to train deep models (Russakovsky et al., 2015).

Researchers have used deep learning to tackle computer vision problems in several areas of civil engineering. Among them, crack detection (Cha et al., 2017; Vu & Duc, 2019; Yeum & Dyke, 2015), structural damage detection (Cha et al., 2018; Gao & Mosalam, 2018; Li et al., 2018; Liang, 2017; Lin et al., 2017), reliability analysis of transportation network (Nabian & Meidani, 2018), traffic congestion and incident detection (Chakraborty et al. 2018b; a), tunnel lining defects detection (Xue & Li, 2018), and pavement crack detection (Zhang et al., 2017) have been investigated by researchers. Image based monitoring and evaluation have also received great attention in the construction engineering domain, due to the dynamic nature and vastness of typical construction sites. For instance, Memarzadeh et al. used handcrafted features, such as Histogram of Oriented Gradients (HOG) and colors, to detect construction equipment and workers (Memarzadeh et al., 2013). Also, Chi and Caldas, proposed a methodology to detect construction vehicles and workers using background subtraction, morphological processing, and neural network for classifying the objects (Chi & Caldas, 2011). Kim et al. proposed a framework to monitor stuck-by accidents using computer vision techniques and fuzzy inference. In the computer vision step, they used background subtraction, morphological operation and object classification and tracking. Afterward, they employed proximity and crowdedness as a contextual construction site information in fuzzy inference step (Kim et al., 2016). However, the abovementioned traditional computer vision solutions inherently suffer from lack of generalization and requires extensive development effort and domain knowledge (LeCun et al., 2015). In contrast, deep learning approach toward computer vision problems introduces an alternative end-to-end solution which is capable of automatic feature extraction without explicit use of domain knowledge based feature selection.

*Arabi@iastate.edu

Among safety specific studies, Fang et al. used Faster-R-CNN (Ren et al., 2015) structure to detect workers and their safety harnesses (Fang et al., 2018b). Faster R-CNN was also used by Fang et al. to detect workers with no hardhat (Fang et al., 2018a). A good comparison between traditional computer vision techniques and deep learning solutions, can be seen by exploring (Fang et al., 2018a) and (Park et al., 2015). Park et al. (Park et al., 2015) employed handcrafted feature extraction, i.e. HOG, to detect human body and hard hat. Then, they used spatial information of hat and body to match detected hats and bodies. Fang et al., however, utilized Faster-R-CNN to directly detect workers with no hardhats. The solution proposed by Fang et al. is end-to-end requiring no handcrafted feature extraction (Fang et al., 2018a). In the activity understanding area, Ding et al. used a hybrid deep learning model to detect dangerous activities. They employed Inseption V3 structure (Szegedy et al., 2015) to extract the features and then used Long Short-Term Memory (LSTM) (Greff et al., 2015) to consider temporal effects and identify unsafe activities (Ding et al., 2018). On a similar note, Luo et al. proposed a convolutional network based solution for recognizing workers' activities. Their study consists of four main steps. First, they track the workers using a single object tracking algorithm. Second, they use optical flow estimation to extract the temporal and spatial information of the objects. Then, they classify the activities in both of the information streams. Finally, they merge the results of temporal and spatial stream classifications to estimate the final activity (Luo et al., 2018b). Luo et al. developed a framework to identify worker activities using still image data. They employed Faster R-CNN to detect the objects and spatial information of the objects in the image to define the activity patterns (Luo et al., 2018a). Construction equipment detection using deep learning has also been conducted by some researchers. Along with using transfer learning (Ling Shao et al., 2015), Kim et al. employed R-FCN (Dai et al., 2016) to detect construction equipment (Kim et al., 2018). Fang et al. used Faster R-CNN to detect workers and excavators in construction sites (Fang et al., 2018c). In a separate study, Son et al., used Faster R-CNN to detect construction workers in various poses and backgrounds (Son et al., 2019). However, almost all of the mentioned endeavors focused on providing solutions disregarding the real-time performance, efficiency, and cost of deployment of the solutions in the construction site.

Based on extensive literature review, we found that most of the studies focus on development of improved techniques for image analytics, but a very few look at the economics of final deployment and the trade-off between accuracy and costs of deployment. In infrastructure management domain we only found one study investigating inference at the edge of the network for road damage detection application (Maeda et al. 2018). However, to the best of our knowledge, there is no such study in construction engineering domain with the focus on the inference using embedded devices. This paper aims at providing the researchers and engineers a practical and comprehensive deep learning based solution to detect construction equipment from the very first step of development to the last step, which is deployment of the solution. The article covers the two phase of practical deep learning solutions. In the first phase, that is the development phase, data gathering and preparation, model selection, model training, and model evaluation are covered. Model optimization, application specific hardware selection, and solution evaluation are conducted in the second phase of the study. The main contributions of this paper can be briefly summarized as follows:

- Improving the AIM construction equipment dataset by adding one more class of equipment and annotating the new images.
- Representing a modified version of SSD-mobilenet object detector which is suitable for embedded systems.
- Proposing several embedded systems for various scenarios in construction engineering domain. Some of these scenarios include but not limited to: (1) ones which require real-time performance such as safety and object tracking; (2) applications which need semi-real-time performance such as productivity analysis, emission analytics, and managerial and security related.

Figure1 shows the general framework, which is followed in this study.

## 2 DEVELOPMENT PHASE

In this section, data gathering and preprocessing and model proposing, training, and evaluation are covered.

### 2.1. Data preparation and labeling

As it is illustrated in Figure 1, this section devotes to the first step of development, which is data gathering and preprocessing. Data can be gathered using three major processes. First, it can be gathered from available large scale datasets, such as ImageNet (Russakovsky et al., 2015), Common Objects in Context (COCO) (Lin et al., 2014), and the Open Image Dataset (Kuznetsova et al., 2018). Second, data can be gathered using web crawling techniques (Olston & Najork, 2010). Finally, image data can be captured at the location of the application by researchers/engineers. First and second approach were used in this study. We used AIM dataset (Kim et al., 2018) which is originally from ImageNet dataset. ImageNet is a large scale dataset and benchmark for computer vision tasks such as classification, detection, and segmentation. It is aimed at covering the majority of the 80000 synsets of WordNet. Currently, it contains more than 14 million images which are hand-annotated (Russakovsky et al., 2015). AIM dataset is a subset of ImageNet which contains construction equipment images, i.e. Excavator, Loader, Roller, Concrete mix truck, and Dump truck. Moreover, to demonstrate a complete data gathering process, web crawling technique is employed to enhance the dataset. Web crawling refers to an automated process in which a crawler (bot) systematically browses the web to retrieve information (Olston & Najork, 2010). Images related to the “Grader” object class were gathered using web crawling.

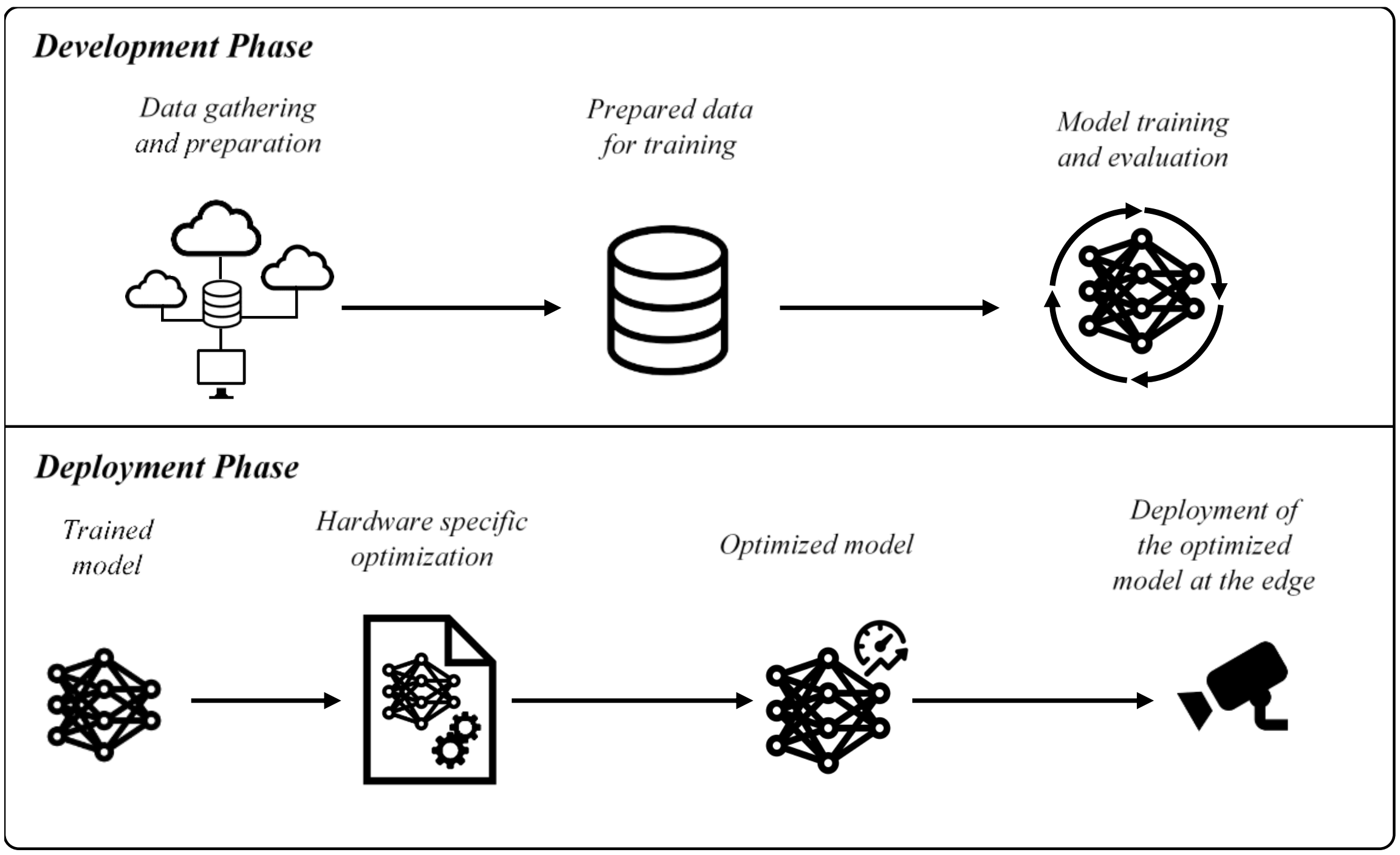

**Figure 1:** The general workflow, followed in this study, as a deep learning based solution for construction equipment detection.

After both visual and automated inspection of the “Grader” images and confirmation of their correctness and quality, they were annotated by the authors. Then, the dataset was split to train, validate, and test datasets. It was ensured that 20% of the initial dataset devotes to test dataset, while 80% of it secured for training and validation. Within the training and validation chuck, 20% of the data devoted to validation and 80% to training. Table 1 summarizes the details of data splitting. A visual inspection of the dataset shows various challenges associated with this computer vision task such as viewpoint variation, scale variation, occlusion, background clutter and intra-class variation. Figure 2 shows some example of these scenarios. Contrary to large scale image datasets, relatively fewer data are available for specialized applications such as detecting construction equipment. Consequently, training a model which is capable of generalization while not underfitted or overfitted, could be unattainable. Moreover, hardware resource restriction in some applications, magnifies the complexity of designing a solution. This is not necessarily the case for other scenarios, such as computer vision competitions. While the authors tried their best to provide comprehensive information about each phase of the study, covering all the technical details are not possible within the paper. However sufficient studies were referred to when other literature covered these details, to understand and guide the process. Subsequent sections detail the procedure of network designing, training, and evaluation.

**Table 1**
Details of train, test, and validation data spliting.

| | Total data | Train | Evaluation | Test |
|---|---|---|---|---|
| **Loader** | 787 | 504 | 126 | 157 |
| **Excavator** | 361 | 231 | 58 | 72 |
| **Dump Truck** | 760 | 486 | 122 | 152 |
| **Concrete mix truck** | 659 | 422 | 105 | 132 |
| **Roller** | 353 | 226 | 56 | 71 |
| **Grader** | 351 | 225 | 56 | 70 |

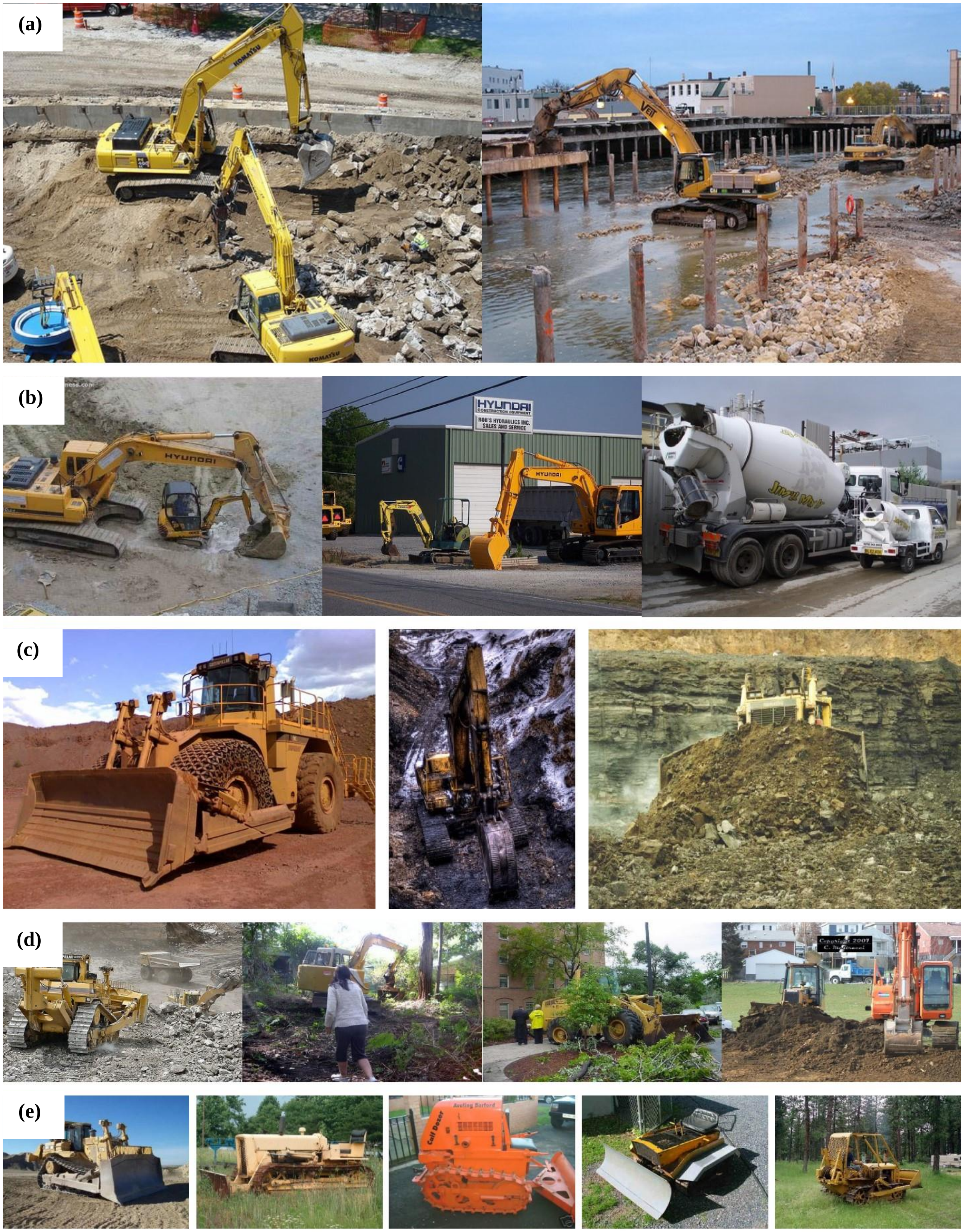

**Figure 2** Examples of challenges associated with visual detection of construction equipements: (a) viewpoint variation, (b) scale variation, (c) background clutter, (d) occlusion, and (e) intra-class variation.

### 2.2. Model training using transfer learning

In this section, feature extractor and detector of the proposed object detection model are described.

#### 2.2.1. Mobilenet as feature extractor

While the general trend in designing deep learning networks is toward deeper models (He et al., 2015; Simonyan & Zisserman, 2014), these models are not optimized for speed and most of them cannot be used for applications which require real-time performance owing to computationally-limited hardware. In contrast, MobileNets and its variants were introduced as an alternative which was optimized primary for speed (Howard et al., 2017).

The main building block of these class of networks are depthwise separable convolutions (Sifre, 2014). Depthwise separable convolution factorize the standard convolution into two distinct operations. In the first operation, separate convolution kernels (also known as depthwise with a convolution) applies to each input channel. Then, pointwise (1 × 1) convolution is used to combine the information of the first operation. Figure 3 illustrates and compares the standard, pointwise and depthwise convolutions. It can be shown that depthwise separable convolutions have less parameter and computational cost than a standard convolution.

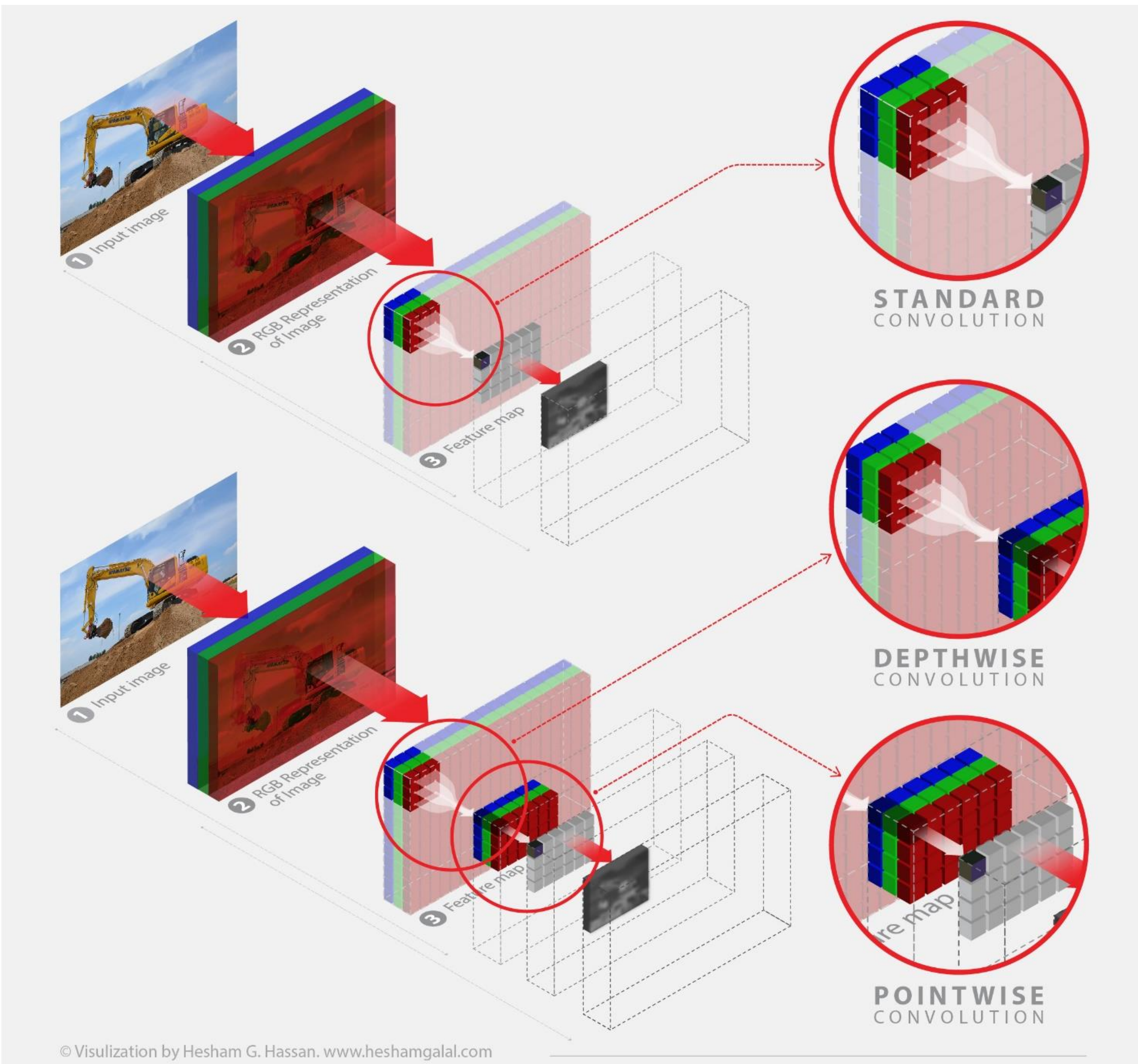

**Figure 3** The infographic of convolution operations.

Table 2 summarizes the parameter size and computational cost of these two types of convolutions. Using Table 2, it can be shown that the reduction in computation and parameter size is $\frac{1}{N} + \frac{1}{D_K^2}$ where $N$ is number of output channels and $D_K^2$ is kernel (filter) size. The depthwise separable convolutions with $3 \times 3$ kernels were used for this study which helped reduce the computation 8 to 9 times compared to standard convolutions. Followed by each convolution in the network, batch normalization (Ioffe & Szegedy, 2015) and Relu6 activation (Krizhevsky, 2016) was used. Figure 4 depicts the depthwise separable convolution block which is used in this study.

**Table 2**
Computation difference between depthwise separable and standard convolution. $N$, $M$, $D_K^2$, and $D_F^2$ are number of output channels, number of input channels, kernel size, and feature map size.

| Convolution | Parameters | Computation Cost |
|---|---|---|
| **Standard** | $D_K^2 \times N \times M$ | $D_K^2 \times D_F^2 \times N \times M$ |
| **Depthwise Separable** | $D_K^2 \times M + N \times M$ | $D_K^2 \times D_F^2 \times M + M \times N \times D_F^2$ |

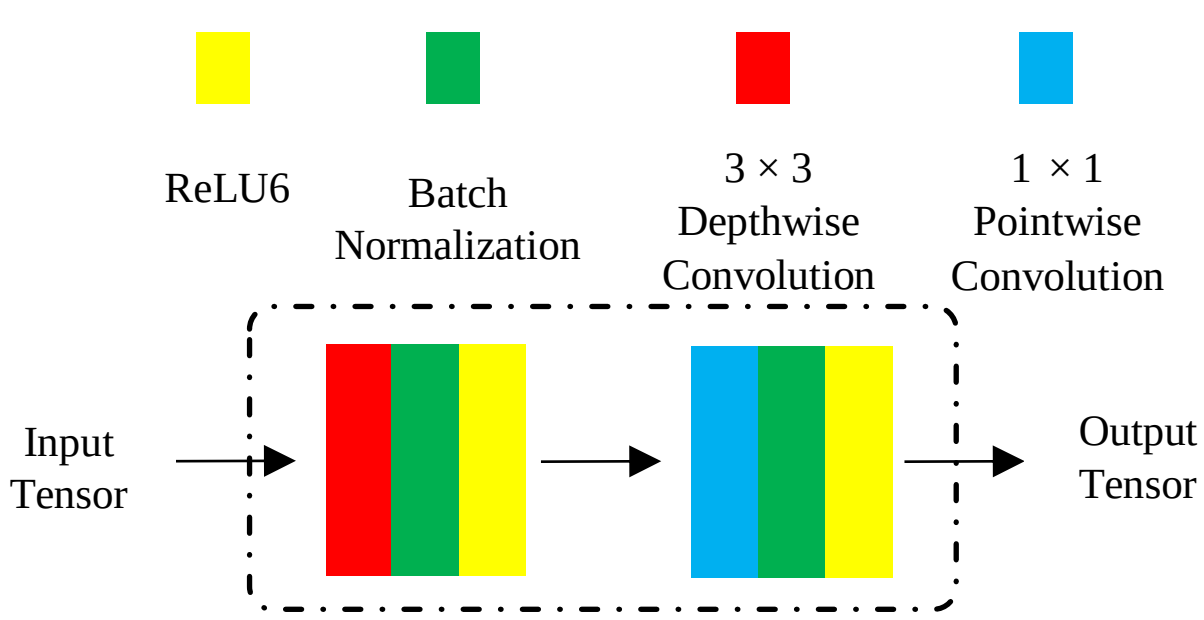

**Figure 4** Depthwise separable convolution block.

### 1.2.2. SSD as the object detector

There are two general types of object detectors i.e. one-stage and two-stage object detectors. In the two-stage object detectors, first, a number of regions with high probability of object existence will be proposed. Then, a convolutional-based classifier will be applied to each proposal. While these types of object detectors are known for their accuracy, they are too computationally intensive to be used with embedded systems and in applications requiring real-time performance (Girshick, 2015; Girshick et al., 2013; Ren et al., 2015). One-stage detectors, however, combine these two steps and perform classification and localization in one single network. End-to-end learning and model simplicity is one of the advantages of the one-stage detectors (Liu et al., 2015; Redmon et al., 2015).

Single Shot Detector (SSD) has been used in this study as the detector. This model uses an auxiliary network for feature extraction, also known as base network. We used mobileNet, previously explained, as the base network here. SSD uses different feature maps- some of them from base network- to perform classification and localization regression. A set of default boxes assigned to each cell of the feature maps. Then, SSD predicts score for each class and four bounding box offsets for each default box at each feature map cell. So, each feature map generates $(c + 4)kmn$ results where $c$ is number of classes, $k$ is number of default bounding boxes and $mn$ is the feature map size. SSD uses multiple feature maps with different size to leverage high level as well as low level information. While the aspect ratio of the default bounding boxes is fixed, the scale of bounding boxes are different at each feature map. Considering employing $m$ feature map in the SSD structure, the scale of the default box $S_k \mid k \in [1, m]$ can be expressed as:

$$S_k = S_{min} + \frac{S_{max} - S_{min}}{m-1}(k-1) \quad (1)$$

where $S_{min}$ is the scale for the lowest feature map and $S_{max}$ is the scale of the highest feature map. In this study, $S_{min}$ and $S_{max}$ was set to 0.2 and 0.9, respectively. The size of the of the default boxes is a function of scale. Five different aspect ratios can be expressed as $a_r \in \{1, 2, 3, 0.5, 0.33\}$ and the width $w$ and height $h$ of each box is $S_k\sqrt{a_r}$ and $S_k/\sqrt{a_r}$, respectively. One more box with scale of $\sqrt{S_k S_{k+1}}$ is also considered for aspect ratio 1. This results in six default box per feature map per cell. The center of each default box is $\left(\frac{i+0.5}{|f_k|}, \frac{j+0.5}{|f_k|}\right)$ where $|f_k|$ is the feature map size and $i, j \in [0, |f_k|]$. SSD uses jaccard overlap (also known as IOU) to identify the matches. Figure 5 illustrates the jaccard overlap calculation. Any default box which has jaccard overlap of 0.5 or greater with ground truth bounding boxes, is considered to be a match.

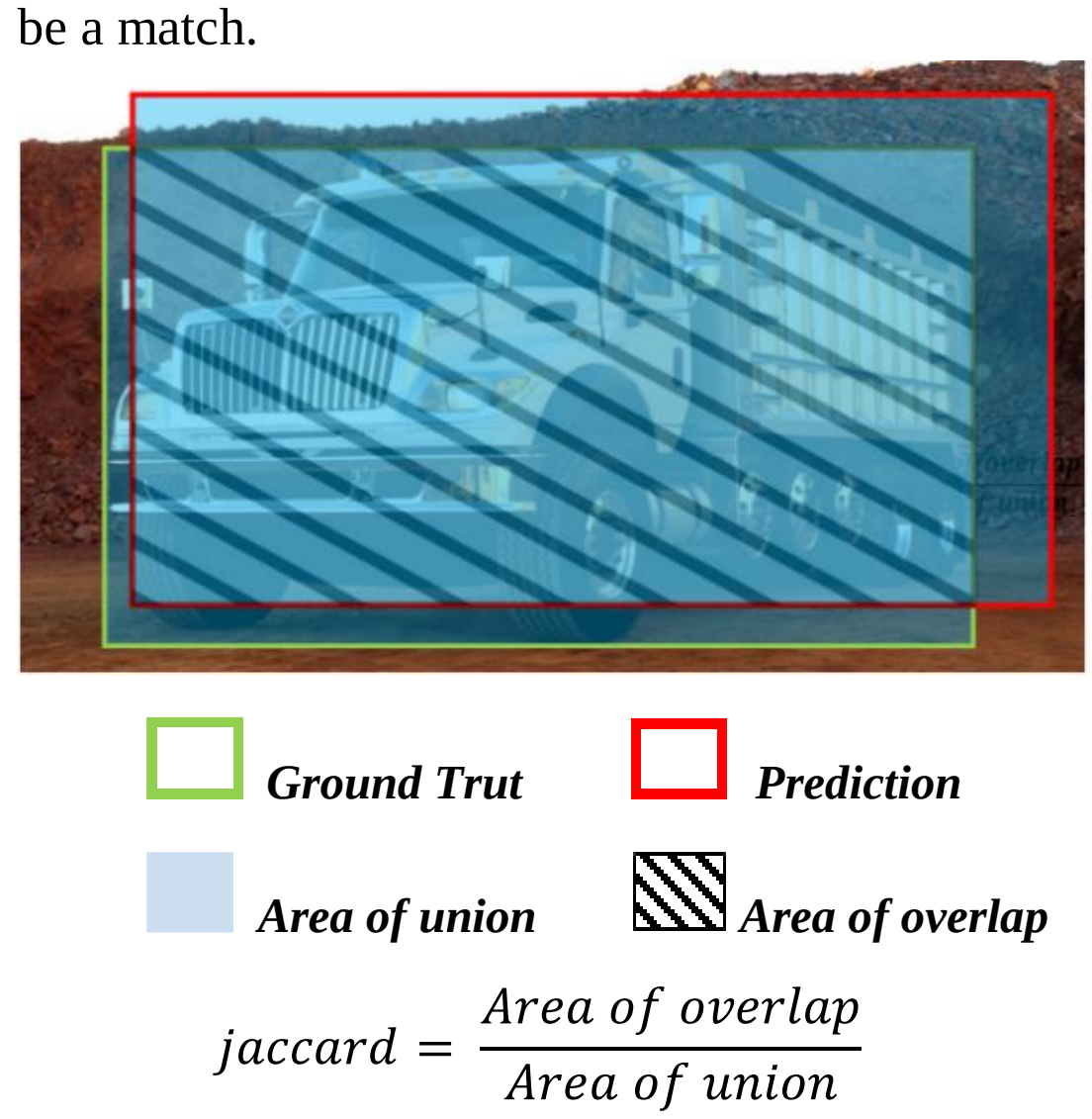

**Figure 5** Illustration of jaccard overlap calculation.

The loss function of the SSD detector can be described as a summation of localization loss, $L_{loc}$, and classification, $L_{conf}$.The localization loss is SmoothL1 loss (Ren et al. 2015):

$$L_{loc} = \sum_{i \in Pos}^{N} \sum_{m \in \{cx, cy, w, h\}} x_{ij}^{k}\ smooth_{L1}\left(l_i^m - \hat{g}_j^m\right) \tag{2}$$

$$\hat{g}_j^{cx} = \frac{g_j^{cx} - d_i^{cx}}{d_i^{w}}\ ,\ \ \hat{g}_j^{cy} = \frac{(g_j^{cy} - d_i^{cy})}{d_i^{h}},\ \hat{g}_j^{w} = \log(\frac{g_j^{w}}{d_i^{w}}),\ \ \hat{g}_j^{h} = \log(\frac{g_j^{h}}{d_i^{h}})$$

where $N$ is the number of positive matches. $l$, $g$, and $d$ are predicted, groundtruth and default bounding boxes. $(cx, cy)$, $w$, and $h$ denotes the center, width, and height of the default bounding box, respectively. An improved version of classification loss is used in this study to boost detector performance. The original SSD model used Hard Negative Mining to address class imbalance issue. After the matching step, a large number of boxes did not have any objects in them and it introduced massive positive and negative imbalances in training dataset. Therefore, a fixed ratio of 3:1 between negative and positive was set to achieve stable and fast training (Liu et al., 2015). However, it was shown that one stage detectors can achieve better performance, even without Hard Negative Mining, by using a different classification loss named focal loss (Lin et al., 2017). This loss adds a modulating factor of $\alpha(1 - c^p)^\gamma$ to the cross entropy loss to down-weight the relative loss of well-classified example and put more focus on a few hard and misclassified examples. Inspired by (Lin et al., 2017), the focal loss is used in this study. So, the classification loss function which was used in this study can be expressed as:

$$L_{conf}(x, c) = -\sum_{i \in Pos}^{N} x_{ij}^{p}\, \alpha\left(1 - \hat{c}_i^p\right)^\gamma \log\left(\hat{c}_i^p\right) - \sum_{i \in N_{eg}} \alpha(1 - \hat{c}_i^0)^\gamma \log(\hat{c}_i^0) \tag{3}$$

where $\hat{c}_i^p = \frac{\exp(c_i^p)}{\sum_p \exp(c_i^p)}$ is predicted confidence for category $p$. $\alpha$=0.75 and $\gamma$=2 was used in this study. The total loss can be summarized as follows:

$$Total\ Loss = \frac{1}{N}(L_{conf} + L_{loc}) \tag{4}$$

where $N$ is the number matched to default bounding boxes. Figure 6 shows the MobileNet-SSD structure. Additionally, Table 3 summarizes the detail information about Mobilenet-SSD structure that was used in this study at layer level.

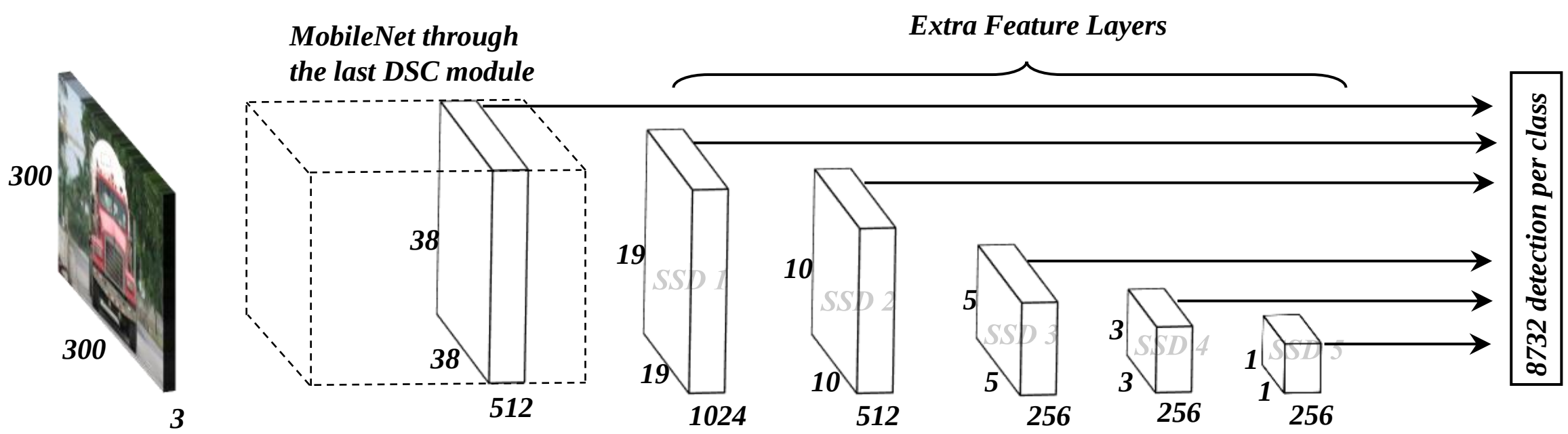

**Figure 6** mobileNet-SSD network architecture.

The Adam (adaptive moment estimation) optimization algorithm is used in order to minimize the loss function and train the network (Kingma & Ba, 2014). This optimizer is an extension of a well-known stochastic gradient descent optimizer, which had shown promising results for solving non-convex optimization problems, in the past. The algorithm uses first moment and second moment of the variable gradient, $g$, to update the variable. The Adam optimization formulation can be summarized as follows:

$$m_t = \beta_1 . m_{t-1} + (1 - \beta_1). g_t \tag{5}$$

$$v_t = \beta_2 . v_{t-1} + (1 - \beta_2). g_t^2 \tag{6}$$

$$\theta_t = \theta_{t-1} - \alpha_t \frac{\frac{m_t}{1 - \beta_1^t}}{\left(\sqrt{\frac{v_t}{1 - \beta_2^t}} + \epsilon\right)} \tag{7}$$

where $\theta$ is the variable that needs to be optimized. $\beta_1$ and $\beta_2$ controls the exponential decay rate of the moments. $\epsilon$ also is a small default number to guarantee the numerical stability of the optimization. 0.9, 0.999, $10^{-8}$ is used for $\beta_1$, $\beta_2$, and $\epsilon$, respectively. Moreover, $\alpha$ denotes learning rate. After comprehensive experiment with different learning rates, an exponentially decayed learning rate was used to train the network. The learning rate formulation can be expressed as:

$$\alpha_t = \alpha_0 * r_d^{\left\lfloor \frac{t}{t_{max}} \right\rfloor} \quad (8)$$

where $\alpha_t$, $\alpha_0$, are $r_d$ is the learning rate at each training iteration, initial learning rate, and decay rate, respectively. $\lfloor \cdot \rfloor$ is floor operation which outputs the closet smallest integer to the input value. $t_{max}$ is set to 1000 in this study.

**Table 3**
MobileNet-SSD structure layers and parameters.

| Type / Stride | Filter Shape | Input Size |
|---|---|---|
| Conv / s2 | 3 × 3 × 3 × 32 | 300 × 300 × 3 |
| Conv dw / s1 | 3 × 3 × 32 | 150 × 150 × 32 |
| Conv / s1 | 1 × 1 × 32 × 64 | 150 × 150 × 32 |
| Conv dw / s2 | 3 × 3 × 64 | 150 × 150 × 64 |
| Conv / s1 | 1 × 1 × 64 × 128 | 75 × 75 × 64 |
| Conv dw / s1 | 3 × 3 × 128 | 75 × 75 × 128 |
| Conv / s1 | 1 × 1 × 128 × 128 | 75 × 75 × 128 |
| Conv dw / s2 | 3 × 3 × 128 | 75 × 75 × 128 |
| Conv / s1 | 1 × 1 × 128 × 256 | 38 × 38 × 128 |
| Conv dw / s1 | 3 × 3 × 256 | 38 × 38 × 256 |
| Conv / s1 | 1 × 1 × 256 × 512 | 38 × 38 × 256 |
| Conv dw / s1 | 3 × 3 × 512 | 38 × 38 × 512 |
| Conv / s1 | 1 × 1 × 512 × 512 | 38 × 38 × 512 |
| 5× Conv dw / s1 | 3 × 3 × 512 | 38 × 38 × 512 |
| 5× Conv / s1 | 1 × 1 × 512 × 512 | 38 × 38 × 512 |
| Conv / s2 | 3 × 3 × 512 × 1024 | 38 × 38 × 512 |
| Conv / s1 | 1 × 1 × 1024 × 1024 | 19 × 19 × 1024 |
| Conv / s1 | 1 × 1 × 1024 × 256 | 19 × 19 × 1024 |
| Conv / s2 | 3 × 3 × 256 × 512 | 19 × 19 × 256 |
| Conv / s1 | 1 × 1 × 512 × 128 | 10 × 10 × 512 |
| Conv / s2 | 3 × 3 × 128 × 256 | 10 × 10 × 128 |
| Conv / s1 | 1 × 1 × 256 × 128 | 5 × 5 × 256 |
| Conv / s2 | 3 × 3 × 128 × 256 | 5 × 5 × 128 |
| Conv / s1 | 1 × 1 × 256 × 128 | 3 × 3 × 256 |
| Conv / s1 | 3 × 3 × 128 × 256 | 3 × 3 × 128 |
| Conv / s1 | 1 × 1 × 256 × 128 | 1 × 1 × 256 |
| Conv / s1 | 3 × 3 × 128 × 256 | 1 × 1 × 128 |

The training process has been carried out by using a high performance GPU (GeForce GTX 1080Ti) and 12 Core i7 3.2 GHz Intel CPUs. Tensorflow-GPU 1.12 and CUDA 9 was used for the training. Transfer learning was used to speed up and enhance the training process. So, the weights were initialized from MobileNet pertained weights value. Initial learning rate was set to $10^{-4}$ and it exponentially decayed during the training by using Equation 8. Figure 7 shows the learning rate variation during the training process. The training loss is illustrated in Figure 8. The minimal variation of loss during training as well as steady convergence to a small number (0.4) shows that the optimizer was able to find the global minimum of the loss function.

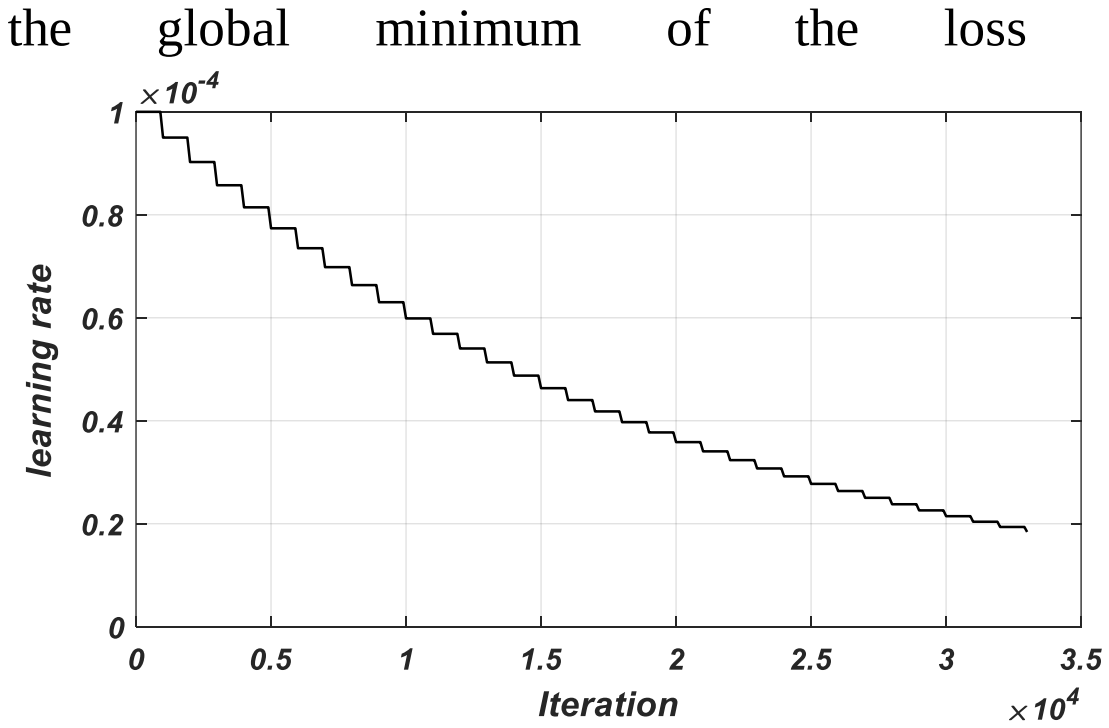

**Figure 7** Training learning rate over training iterations.

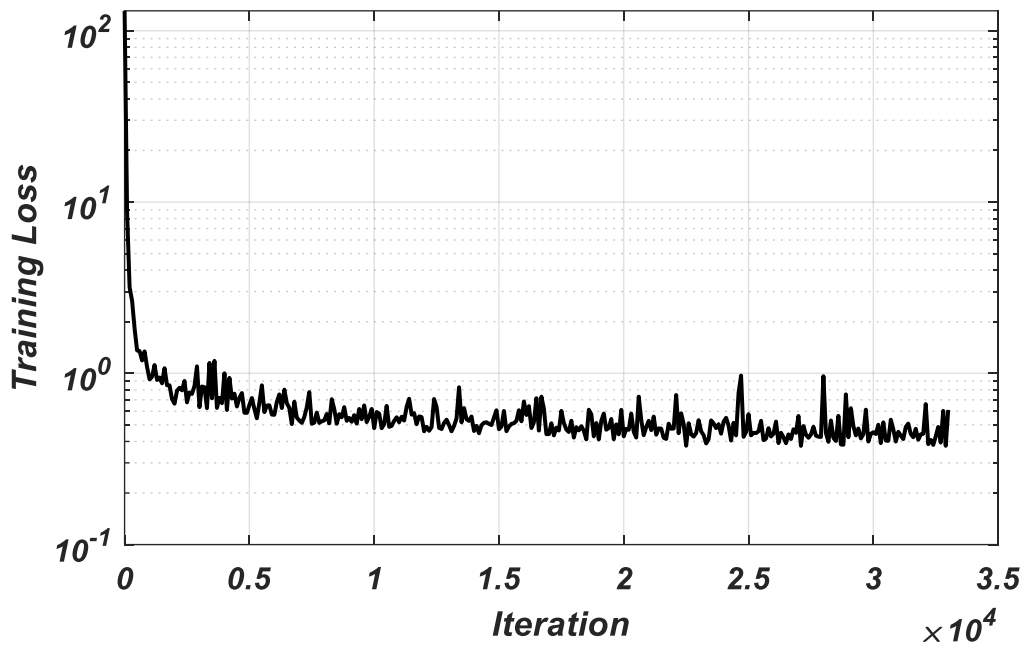

**Figure 8** Training loss over training iterations.

### 2.2.3. Model Evaluation and metrics

The PASCAL VOC challenge evaluation metric is used to evaluate the localization and classification performance of the detector (Everingham et al., 2010). The output of the detector model for each image is a set of bounding boxes and their associated confidence score indicating the probability of belonging to a certain class. Classification is evaluated by identifying weather the predicted class is identical with the groundtruth class while localization is measured by using the jaccard overlap (Figure 5). Based on the PASCAL VOC metrics, any detection with jaccard overlap ≥ 50% and identical prediction label with groundtruth, is considered as a correct detection. The Precision and Recall are the main building blocks of this evaluation and can be expressed as follows:

$$Precision = \frac{TP}{TP + FP} \quad (9)$$

$$Recall = \frac{TP}{TP + FN} \quad (10)$$

where $TP$, $FP$ and $FN$ are true positive, false positive and false negative, respectively. Intuitively, Precision measures how accurate the detection is, while Recall measures how complete the detection is; So the ideal detector is the one which has the highest accuracy (*Precision=1*) and completeness (*Recall=1*). The value of $FP$, $FN$ and $TP$ can be changed by setting different threshold values for confidence score. Consequently, it is important to evaluate Precision and Recall at different thresholds to measure the overall performance of the detector model for each category. Interpolated Average Precision (AP) could be defined as the average of maximum precision at different recalls and can be defined as:

$$AP = \frac{1}{11} \sum_{r \in \{0.0,\ldots,1.0\}} \max_{\tilde{r} \geq r} p(\tilde{r}) \quad (11)$$

where $p(r)$ is Precision at the Recall $r$. mAP is also defined as the average of APs over all the categories. By using the test dataset, AP was calculated for each category. The AP is identical with the area under the Precision-Recall curves which are illustrated in Figure 9. Table 4 summarizes the details of model evaluation for all the categories. To validate and monitor the performance of the detector, mAP and loss was calculated by using the validation dataset during the training process. This helped to ensure that the detector maintains the overall performance during training without overfitting or losing its generalization. A review of Figure 10 and Table 4 indicates that validation mAP is very close to the test mAP.

**Table 4**

Average precision (AP) for each object category derived using training model and test dataset.

| | Dump truck | Excavator | Grader | Loader | Mixer truck | Roller | |
|---|---|---|---|---|---|---|---|
| AP | 92.31% | 83.70% | 93.86% | 93.77% | 96.94% | 86.65% | mAP=91.20% |

As demonstrated in the previous section of the study, the proposed model was able to detect the majority of the objects with high precision and recall. Figure 11 shows the results of the detection for various categories. However, it is imperfect in some hard cases. Figure 12 depicts detection failure cases. Misclassified, merged and missed detections is among the unsuccessful detections. The poor performance in these cases is mainly due to the relatively small number of training images. Having more training data with various conditions such as, different orientation, scale, location and brightness, will definitely improve the detection performance. Once the development phase is completed, the model is ready to be optimized and deployed at the location of the application. The subsequent section devotes to deployment phase.

## 3 DEPLOYMENT PHASE

The paradigm of inference at edge along with related embedded systems are covered in the subsequent sections. Two separate embedded systems were proposed to address the needs of two distinct scenarios. Nvidia Jetson TX2 along with TensorRT optimization is introduced as a GPU accelerated solution for the applications which needs a real-time, yet accurate performance. Additionally, Rassbperry Pi 3B+ along with Intel Neural Computing Sticks (NCS) was proposed for low demand applications.

### 3.1. Inference at the edge

The availability of labeled data, generated by various types of sensors and devices, together with the recent progresses in the artificial intelligence area, introduces innovative applications such as Connected Autonomous Vehicles (CAVs), smart cities, and Intelligent Infrastructures (IEs). There are two approaches for intelligent decision-making, namely, cloud computing based decision making, and edge computing based decision making. Cloud computing refers to a set of computing services such as, servers, storage, analytics, databases etc., which are delivered over the internet. In this model, data acquisition is conducted at the edge of a network (sensors), then, the data is sent to the cloud for processing and decision making. While this solution is relatively fast and easy to set up, it is associated with some inherent limitations, some such examples being latency and jitter, limited bandwidth, and personal data privacy and security (Ericsson AB, 2016). On the other hand, in the edge computing paradigm, gathering, storing, processing, and decision making can all be done at the edge of a network. Several benefits can be expressed for edge computing (Edge intelligence, 2016):

- Efficient and fast intelligent decision-making by deploying machine learning algorithm at the edge of the network, thereby eliminating the roundtrip delay introduced by cloud computing paradigm.
- Securing data close to its origin and being able to follow local management and control policies.
- Fast recovery from network failure or maintenance.
- Decreasing the data transfer cost by lowering communication over public network. Only alarms or decisions can be sent to the cloud servers.

Two edge computing platform have been introduced in this study, i.e. Raspberry Pi3 (R. PI.) with intel neural compute stick and Nvidia Jetson TX2.

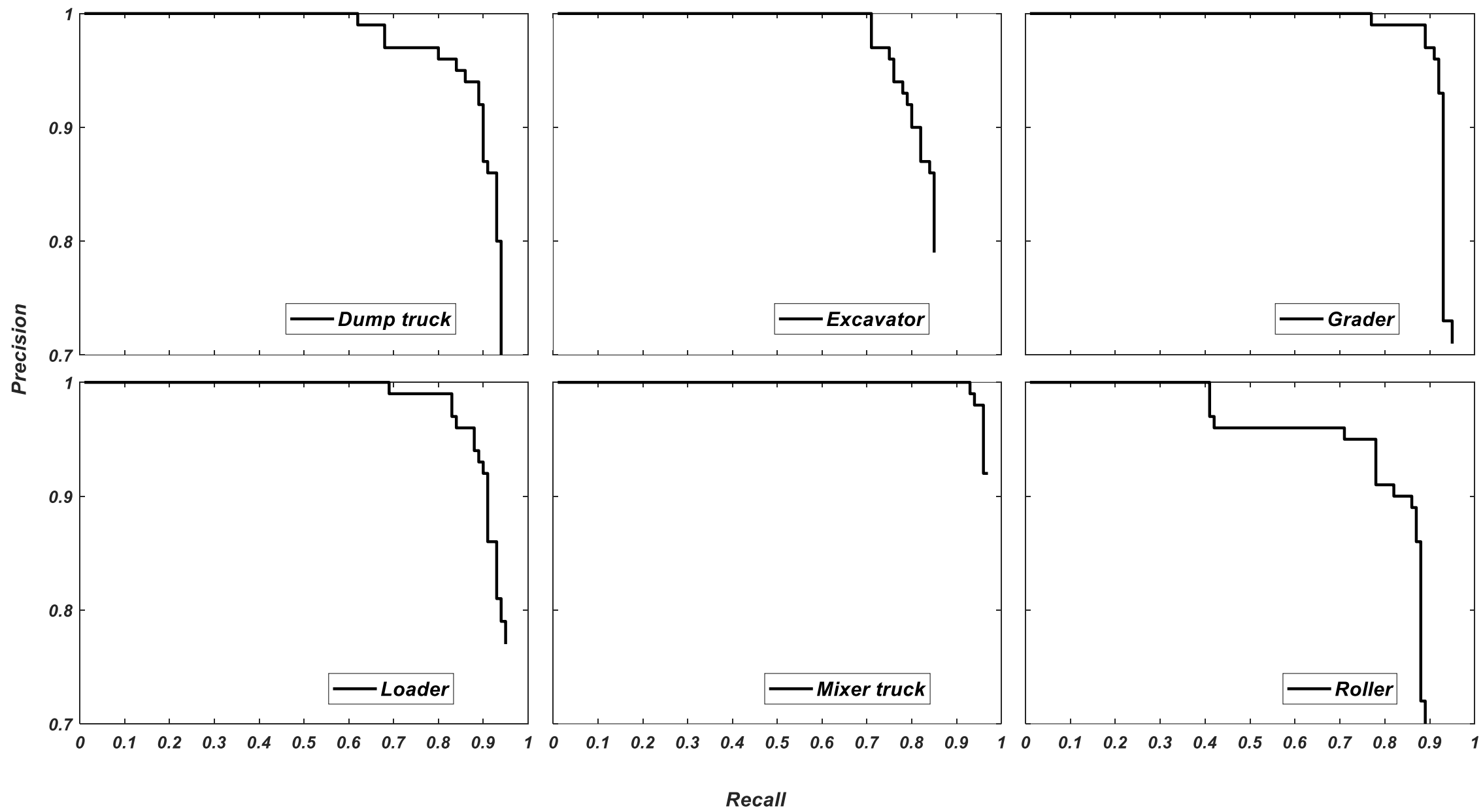

**Figure 9** Precision-Recall curves derived using test dataset for various object categories.

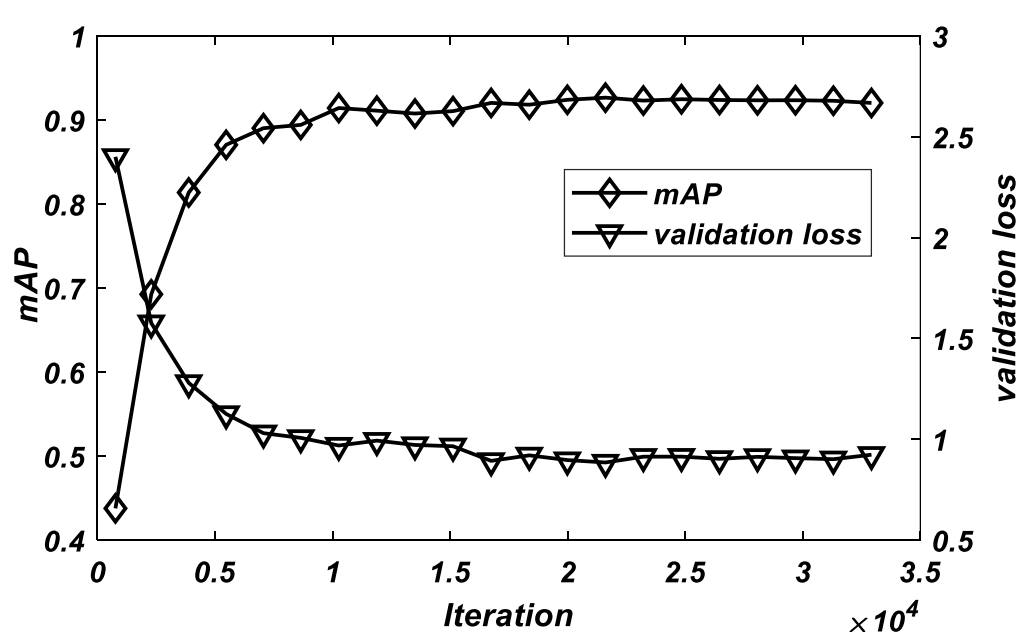

**Figure 10** Variation of validation loss and mAP over training iterations.

#### 3.1.1. Jetson TX2

Nvidia TX2 uses Tegra system-on-chip (SoC) and has the size of a credit-card with input, output and processing hardware, similar to a typical computer. It takes advantage of Nvidia GPUs which enables it to accelerate deep learning related computations. The width and height of the TX2 is 50mm and 87mm, respectively. Table 5 summarizes the technical specifications of TX2. TX2 module is called Jetson TX2 development kit when it is mounted on a 7" × 7" printed circuit which contains typical input and output ports. Figure 13 shows a view of Jetson TX2. The neural network can be trained by using a host machine with a powerful GPU or by using a gpu-enabled cloud compute instances. Then it can be optimized and deployed on the TX2 module. Nvidia Jetpack should be used as the Software Development Kit (SDK) for Nvidia Jetson TX2. In this study, Jetpack 3.2 is used to flash the TX2. Jetpack should be installed on the host machine as well as the Nvidia TX2. By doing so, necessary toolkits and packages such as CUDA, CUdnn and TensorRT get installed on TX2. CUDA is a parallel computing platform which increases the computing performance by utilizing GPUs. CUDA deep neural network (cuDNN) is a gpu-accelerated library which includes highly tuned implementation of operations such as convolution, pooling and activation. TensorRT is a platform for high-performance deep learning Inference which includes optimizer and runtime, enabled to make application with low-latency and high-throughput. TensorRT is a C++ library which improves the inference performance on NVIDIA GPUs. The Input of the TensorRT optimizer is a trained neural network and its output is an optimized inference engine. The inference engine is the only thing that needs to be deployed in the production environment. TensorRT enhances latency, power and memory consumption, and the throughput of the network by combining layers and optimizing the kernel selection. It can further improve the network performance by running it in lower precision. For instance, it eliminates the layer whose outputs are not used, horizontal and vertical fusion of convolution and activation operations, and adjusting the precision of weights from FP32 to FP16 or INT8. Figure 14 summarizes the TensorRT workflow. The performance of the optimized model is discussed in the subsequent section.

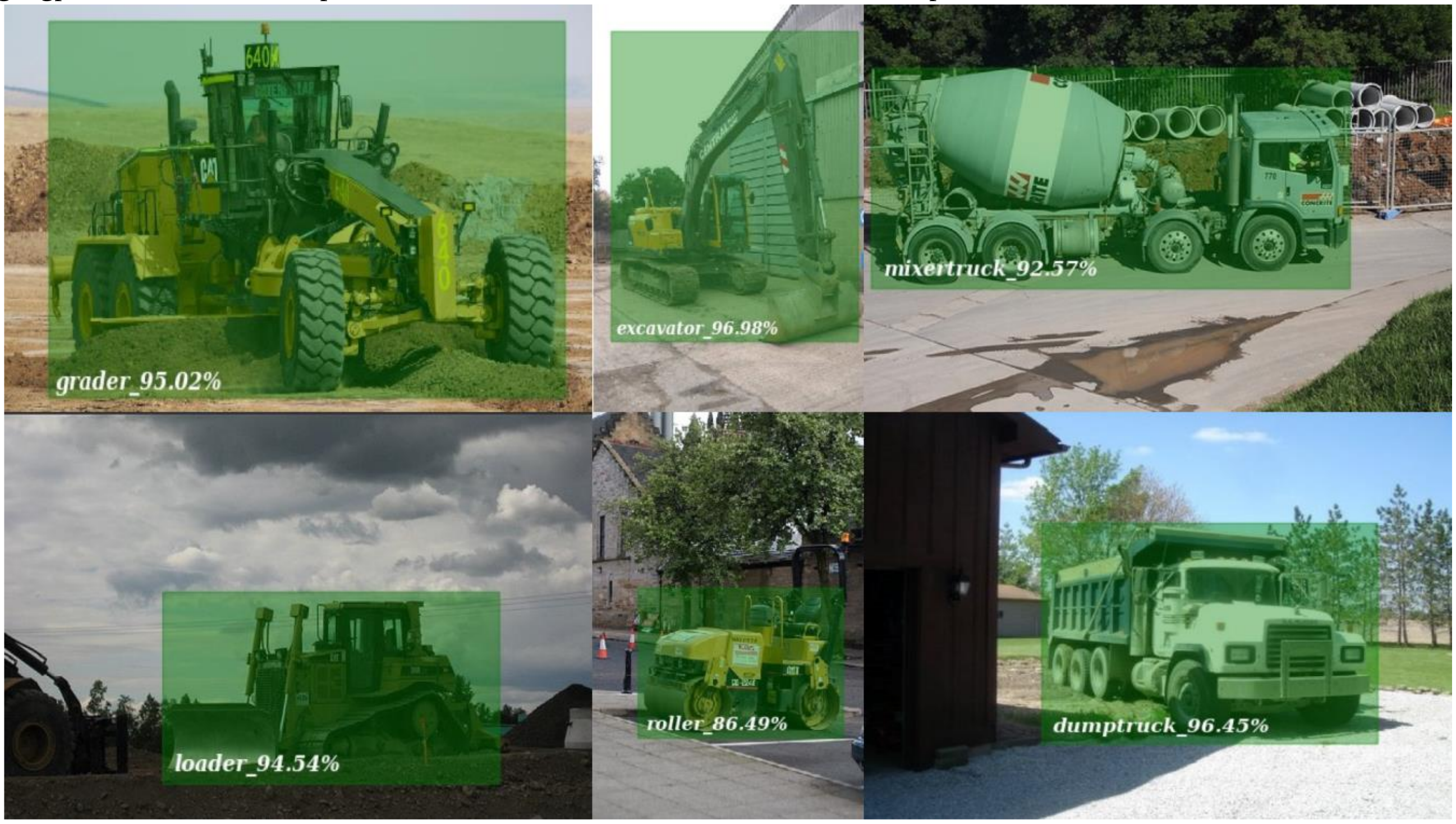

**Figure 11** Examples of successful detection with proposed MobileNet-SSD structure.

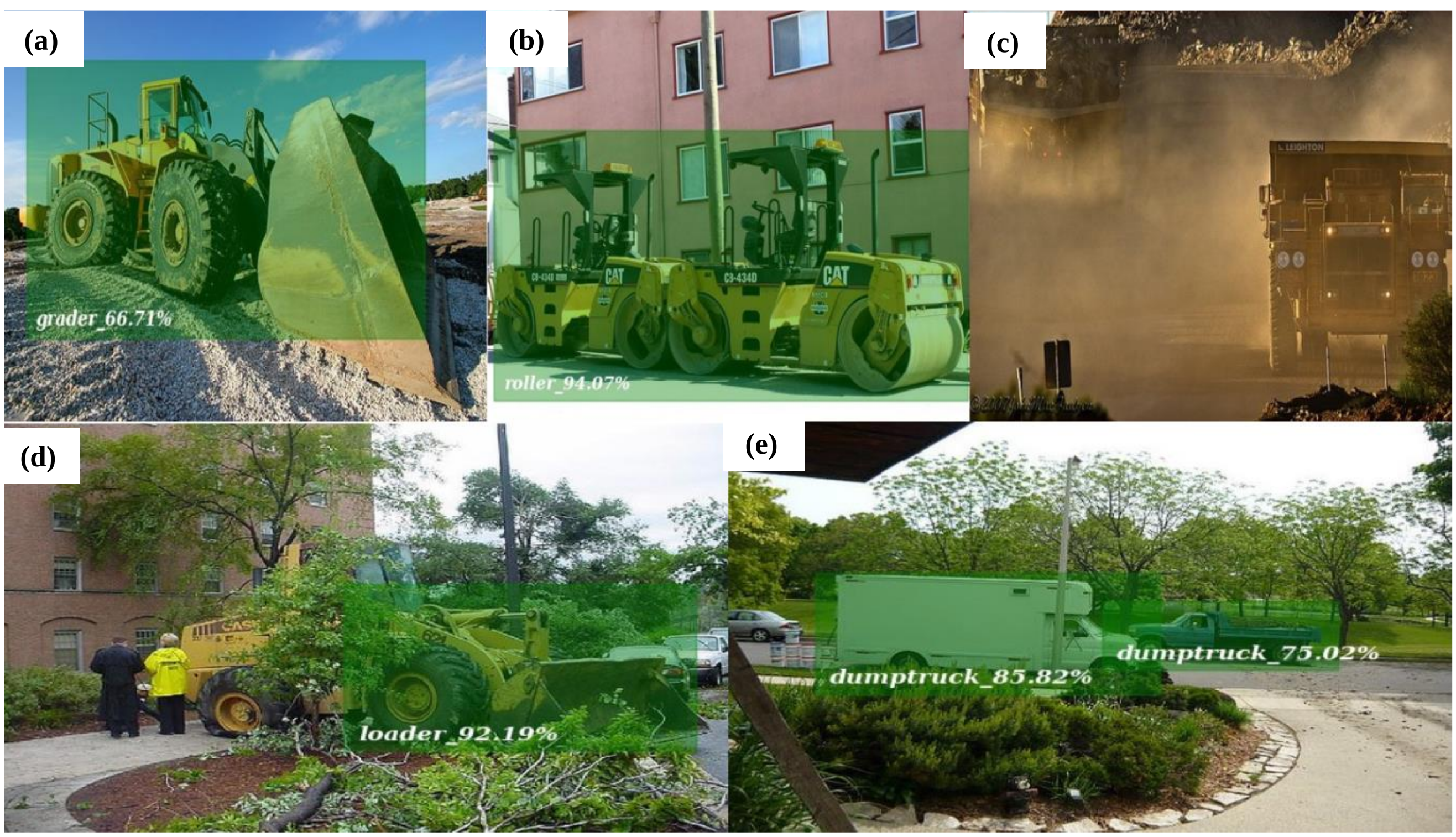

**Figure 12** Examples of failure in detection: (a) misclassified, (b) merged, (c and d) missed, and (e) wrong classification.

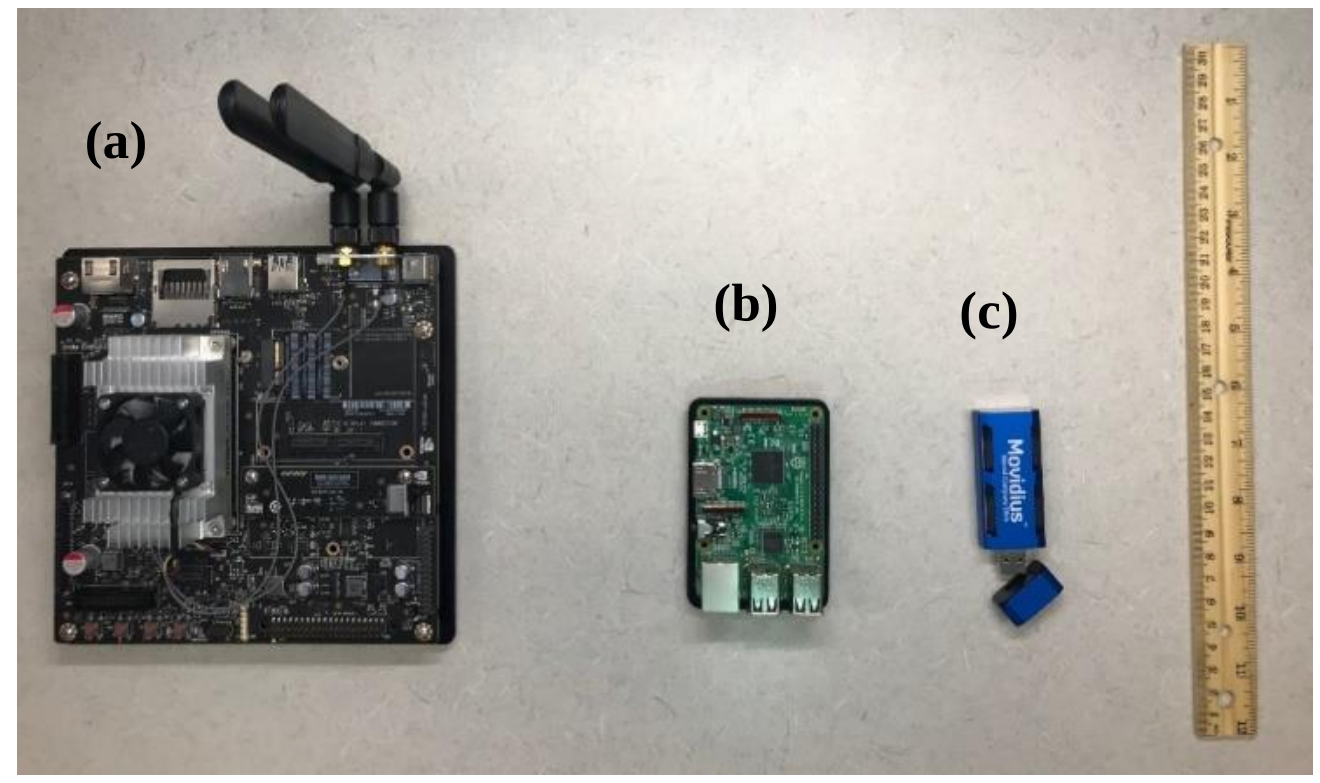

**Figure 13** A view of (a) Jetson TX2, (b) Raspberry Pi 3B+, and (c) intel NCS.

**Table 5** Detailed specification of Jetson TX2, Raspberry Pi 3B+, and Intel NCS.

| | **Jetson TX2** | **Raspberry Pi 3 B+** | **Intel NCS** |
|---|---|---|---|
| **GPU** | NVIDIA Pascal, 256 CUDA cores | Broadcom VideoCore IV | Intel® Movidius™ Myriad™ 2 Vision Processing Unit (VPU) |
| **CPU** | HMP Dual Denver 2/2 MB L2 + Quad ARM® A57/2 MB L2 | 4× ARM Cortex-A53, 1.2GHz | N.A. |
| **Memory** | 8 GB 128 bit LPDDR4 59.7 GB/s | 1GB LPDDR2 (900 MHz) | N.A. |
| **Display** | 2x DSI, 2x DP 1.2, HDMI 2.0, eDP 1.4 | HDMI, DSI | N.A. |
| **Data storage** | 32 GB eMMC, SDIO, SATA | microSD | N.A. |
| **USB** | USB 3, USB 2 | USB 2 | N.A. |
| **Connectivity** | 1 Gigabit Ethernet, 802.11ac WLAN, Bluetooth | 100 Base Ethernet, 2.4GHz 802.11n wireless | USB 3 |
| **Mechanical** | 50 mm × 87 mm | 56.5 mm × 85.60 mm | 72.5 mm x 27 mm |

### 3.1.2. Raspberry Pi and Intel NCS

The Raspberry Pi3 Model B+ is the latest version of the Raspberry Pi which uses a SoC, the size of a credit card and can function similar to a standard computer capable of high performance in basic computer tasks. Its low cost and tiny size made it ideal for embedded systems in particular [55]. Table 5 summarizes the main technical specifications of R. Pi 3B+. While R. Pi might be suitable for basic computer tasks, it cannot deliver high performance for computationally intensive tasks like object detection. We added Intel NCS as a deep learning accelerator to the proposed system. NCS is a USB-drive-sized fan-less deep learning device which can accelerate computationally intensive inference, at the edge. This device is powered by an Intel Movidious Vision Processing Unit (VPU) which optimizes the neural network operations. It is an ideal compact deep learning inference accelerator for resource restricted platforms, such as R. Pi. It supports Tensorflow [56] and Caffe [57] deep learning frameworks. Detailed technical specifications of the NSC can be found in Table 5. NCS consumes only 1 W of power and has proven that it can highly speedup inference over Raspberry Pi CPU (Intel® Movidius™ Neural Compute Stick | Intel® Software, 2019). In order to make the mentioned setup functional, a trained neural network should be first converted into an Intermediate Representation (IR) using the OpenVINO toolkit, provided by Intel. Then, the optimized IR can be used for inference. Figure 13 shows a view of the Raspberry Pi and NCS.

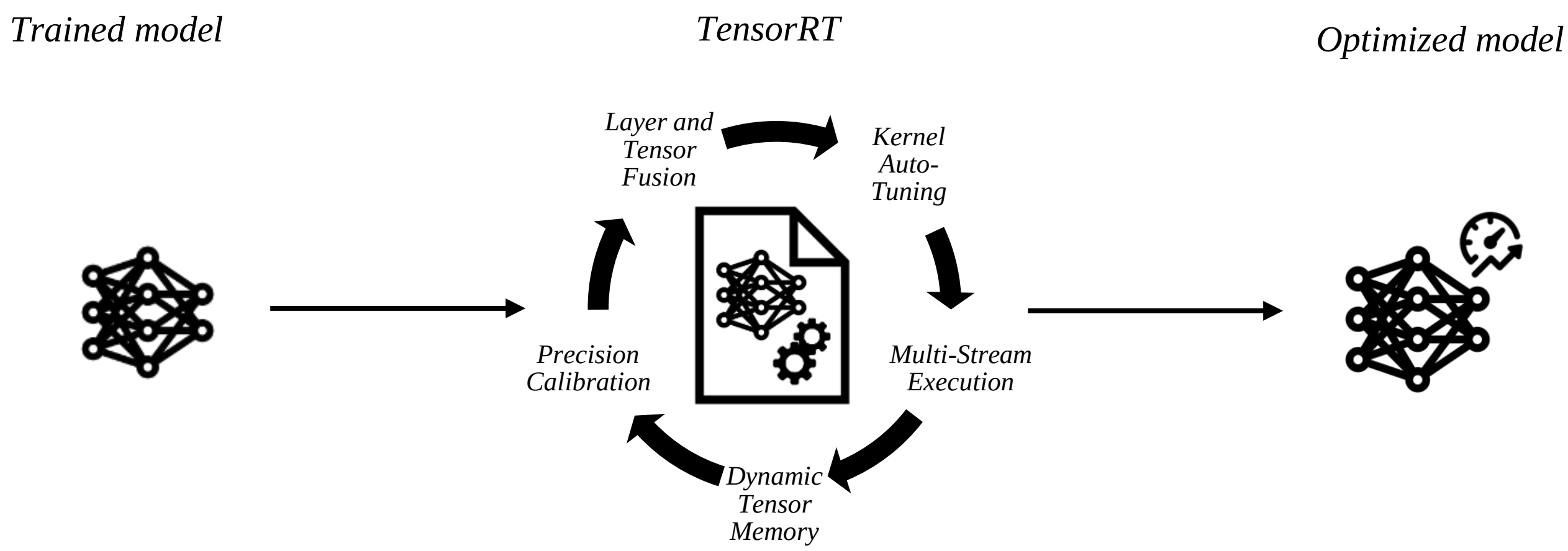

**Figure 14** Nvidia TernsorRT optimization frame work

## 4 RESULT AND DISCUSSION

A trained neural network can be deployed on either cloud or on embedded systems. As construction is typically a long-term process, utilizing cloud services is fairly expensive. For example, *Amazon Machine Learning* will cost more than $90 for 20 hours of compute time and 890000 batch predictions. Consequently, we mainly focused on the embedded systems in the context of the current study.

Inference speed, efficiency, and normalized benefit of four setups were investigated for several embedded systems, i.e. (1) a jetson TX2 without TensorRT optimizations, (2) a jetson TX2 with TensorRT optimizations, (3) a Rassbpery Pi 3B+ with Intel NCS, and (4) inference with GTX 1080 Ti with Intel Core i7. It should be noted that all the benchmarking was done by operating the Jestson TX2 on maximum performance. Reviewing the results for the first setup shows that the inference speed is 25 frame per second (FPS). In order to compare the inference accuracy of this setup with others, AP for each category was calculated using the test dataset and mAP=93.41% was achieved. In the second setup, the neural network was optimized for Jetson TX2 Using TensorRT. Examination of the optimized model showed that it is able to obtain the inference speed of 47 FPS, which is well above the inference speed needed for real time applications. Figure 15 (a) demonstrates a detection result along with inference speed examination, using Jetson TX2 and TensorRT. TensorRT is able to speed up the inference speed with the cost of reduction in inference accuracy. Half precision floating point (FP16) accuracy has been used in this study. Great inference speed up (25 FPS to 47 FPS) achieved with the cost of minimal mAP reduction (93.41% to 91.36%) by utilizing TensorRT. This setup is especially ideal for safety as well as object tracking application which require real-time process.

As elaborated in section 2.1.2, an Intermediate Representative (IR) is needed to conduct inference with the R. Pi and NCS setup. It should be noted that NCS is not necessary in this setup. Inference can be conducted using Opencv with IR backend on R. Pi. However, the inference speed is very low duo to R. Pi limited computational performance (about 0.25 FPS was achieved in this study). 8 FPS was achieved by adding the NCS to the R. Pi setup. Adding an NCS to the setup increased the inference speed by 32 times. Additionally, inference accuracy (mAP) of this setup is 91.22 %. It should be noted that multiple NCSs can be used together to further enhance the inference speed. A detailed inference accuracy comparison between proposed embedded systems can be found in Table 6. Figure 15 (b) illustrates a detection result using R. Pi with Intel NCS. This setup is particularly suitable for any application which needs semi-real-time performance. For example, semi-real-time tracking of construction equipment might be of interest for productivity and emission analytics. Moreover, for managerial and security purposes, this setup can be used as a video recording trigger in certain situations. This can save several storage spaces and facilitate the inspection process.

A Desktop PC can also be used as an inference system at the edge. However, this system requires Fiber Optic Cable which is very expensive. If a construction site already had fiber cables for other purposes, this method can be adapted. With the setup mentioned in the current study, one stream of video can be processed with an inference speed of 166 FPS and mAP of 91.20%. Since different Tensorflow versions were used to generate .pb frozen graph in this setup and that of Jetson TX2 without TensorRT optimization, a slight inconsistency between mAPs is anticipated.

Figure 16 (a) compares the inference speed of the mentioned four setups. Additionally, inference efficiency was also investigated for each setup. Inference efficiency can be measured by dividing inference speed by power consumption, namely, $FPS/Watt$. Jetson TX2 consume 15 W of power at maximum performance, R. Pi 3B+ and Intel NSC consume 6 W, and desktop PC (GTX 1080 GPU with Intel Core i7 CPU) is estimated to consume almost 850W. Figure 16 (b) summarizes the inference efficiency of different setups. Moreover, normalized inference benefit analysis was conducted for the proposed embedded systems. The price of the development kit was considered for this analysis. The price of Jetson TX2, R. Pi, and Intel NCS is $600, $75, and $75, respectively at the time of writing this article. The desktop PC cost is roughly $1700. Figure 17 (c) depicts the normalized inference benefit of the studied systems. Based on the results of the conducted analysis R. Pi with NCS has the highest inference benefit.

## 5 CONCLUSION

A comprehensive deep learning based solution for construction equipment detection was proposed in this study. While the focus was on the deployment of the solution in this work, its development was also covered. In development phase, first, available labeled datasets and web crawling technique was used for data gathering. Then, a modified version of MobileNet-SSD was proposed as the object detection model. The model was carefully selected to fit the hardware-restricted nature of the embedded systems. The trained model was evaluated and its generalization was ensured. Afterward, two main separate embedded systems were proposed to address the needs of several scenarios. Nvidia Jetson TX2 along with TensorRT optimization was introduced as a GPU accelerated solution for the scenarios which needs a real-time yet accurate performance such as safety and construction equipment tracking applications.

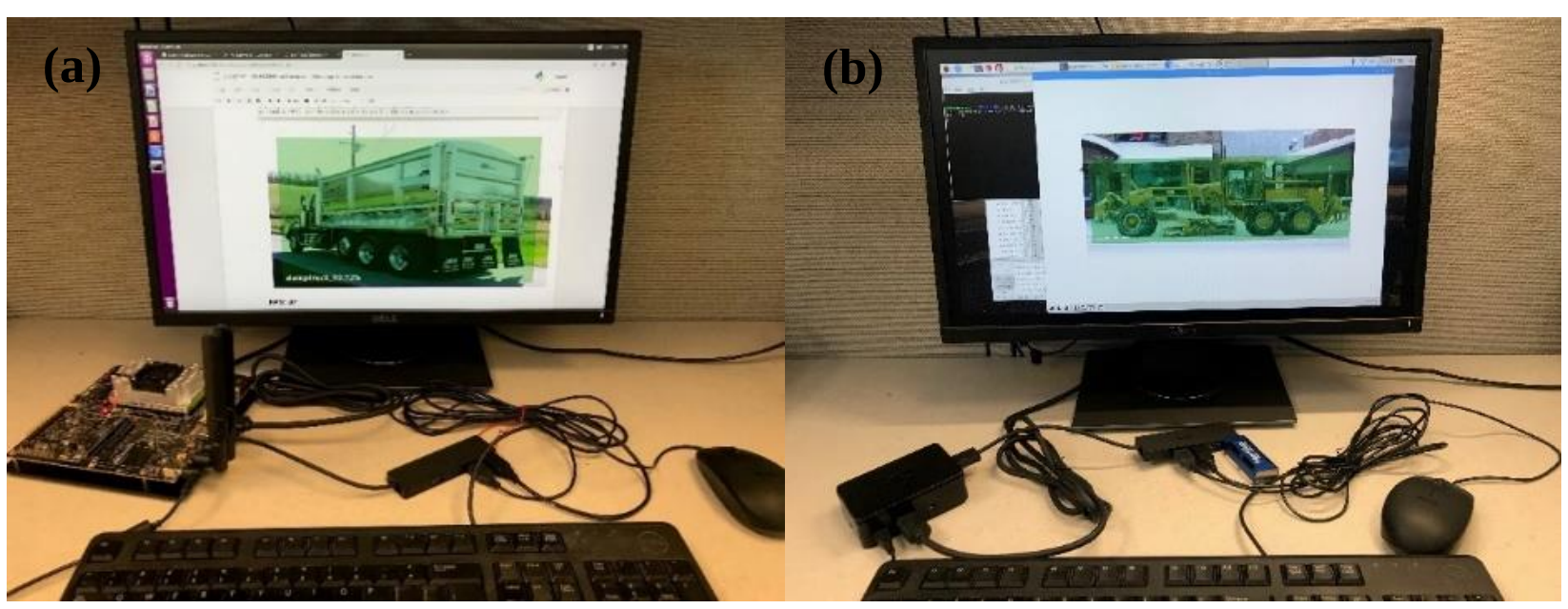

**Figure 15** Demonstration of deployment of the optimized model on the (a) Nvidia Jetson TX2 and (b) Raspberry Pi 3B+ with intel NCS.

**Table 6** AP for each object category derived using the optimized models on embeded systems.

| | **Dump truck** | **Excavator** | **Grader** | **Loader** | **Mixer truck** | **Roller** | |
|---|---|---|---|---|---|---|---|
| **GTX 1080 with Intel Core i7 [$1700]** | 92.31% | 83.70% | 93.86% | 93.77% | 96.94% | 86.65% | mAP=91.20% |
| **Jetson TX2 without TensorRT [$600]** | 93.73% | 87.74% | 94.28% | 96.43% | 98.49% | 89.78% | mAP=93.41% |
| **Jetson TX2 with TensorRT [$600]** | 92.29% | 83.67% | 93.86% | 93.77% | 96.95% | 87.63% | mAP=91.36% |
| **Raspberry Pi 3 B+ with NCS [$150]** | 92.30% | 83.73% | 93.87% | 93.79% | 96.95% | 86.70% | mAP=91.22% |

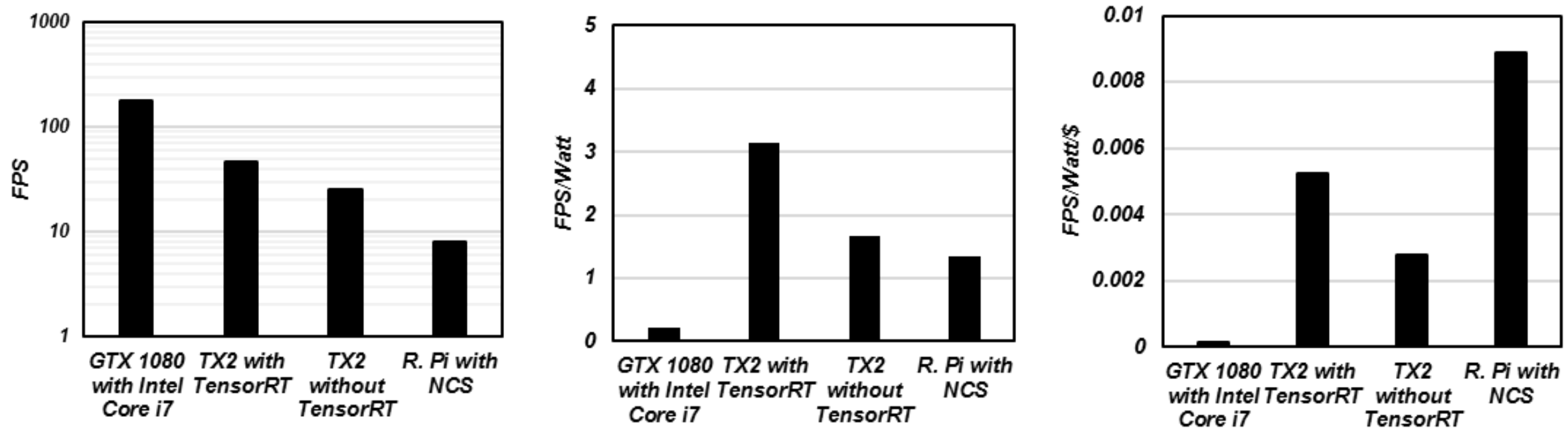

**Figure 16:** Inference speed, eficiency, and normalized benefit comparison of proposed embeded systems.

91.36% mAP and 47 FPS were achieved by evaluating the Nvidia Jetson TX2 with TensorRT optimizations. Moreover, Rassbperry Pi 3B+ with Intel NCS was proposed for low demand applications. mAP of 91.22% and 8 FPS was obtained from this embedded system. This setup is particularly suitable for scenarios requiring semi-real-time performance such as productivity and managerial related applications. Among the proposed embeded systems, Jetson TX2 with TensorRT optimizations has the highest inference speed and efficency, and Rasbperry Pi 3B with Intel nural stick associated with the highest normalized inference benefit. The outcome of this study can be used for several purposes such as, safety monitoring, productivity assessments, and managerial decisions.